\documentclass{article}

\usepackage{PRIMEarxiv}

\usepackage[utf8]{inputenc} 
\usepackage[T1]{fontenc}    
\usepackage{hyperref}       
\usepackage{url}            
\usepackage{booktabs}       
\usepackage{amsfonts}       
\usepackage{nicefrac}       
\usepackage{microtype}      
\usepackage{lipsum}
\usepackage{graphicx}
\usepackage{xcolor}
\graphicspath{{media/}}     
\usepackage{amsmath} 
\usepackage{amssymb} 
\usepackage[table]{xcolor}
\usepackage{multirow}   
\usepackage{multicol}
\usepackage{setspace}   
\usepackage[raster,skins, most]{tcolorbox} %
\usepackage{pdfcol}
\tcbuselibrary{breakable}
\newcommand{\uag}[1]{\textcolor{green!60!black}{\small(+#1)}} 
\newcommand{\dab}[1]{\textcolor{red!80!black}{\small(-#1)}} 
\definecolor{blue1}{RGB}{66, 121, 163}
\definecolor{green1}{RGB}{113, 126, 108}
\definecolor{gray1}{RGB}{83, 82, 82}

\usepackage[most]{tcolorbox}
\usepackage{listings}
\usepackage{xcolor}

\lstdefinelanguage{json}{
    basicstyle=\small\ttfamily,
    breaklines=true,            
    breakatwhitespace=false,    
    columns=flexible,
    keepspaces=true,
    stringstyle=\color{blue!70!black},
}

\title{Yunque DeepResearch Technical Report
}

\author{
Yuxuan Cai$^1$, Xinyi Lai$^1$, Peng Yuan$^{1,2}$, Weiting Liu$^{1,3}$, Huajian Li$^1$, Mingda Li$^1$, \\ [5pt] \textbf{Xinghua Wang$^1$, Shengxie Zheng$^1$, Yanchao Hao$^1$, Yuyang Yin$^{1*}$, Zheng Wei$^{1*}$} \\
\\
$^1$Tencent BAC $\quad$
$^2$Tsinghua University $\quad$
$^3$Fudan University
}

\begin{document}
\maketitle

\begin{abstract}

Deep research has emerged as a transformative capability for autonomous agents, empowering Large Language Models to navigate complex, open-ended tasks. However, realizing its full potential is hindered by critical limitations, including escalating contextual noise in long-horizon tasks, fragility leading to cascading errors, and a lack of modular extensibility. To address these challenges, we introduce \textbf{Yunque DeepResearch}, a hierarchical, modular, and robust framework. The architecture is characterized by three key components: (1) a centralized \textit{Multi-Agent Orchestration System} that routes subtasks to an \textit{Atomic Capability Pool} of tools and specialized sub-agents; (2) a \textit{Dynamic Context Management} mechanism that structures completed sub-goals into semantic summaries to mitigate information overload; and (3) a proactive \textit{Supervisor Module} that ensures resilience through active anomaly detection and context pruning. Yunque DeepResearch achieves state-of-the-art performance across a range of agentic deep research benchmarks, including GAIA, BrowseComp, BrowseComp-ZH, and Humanity’s Last Exam. We open-source the framework, reproducible implementations, and application cases to empower the community.
{\let\thefootnote\relax\footnotetext{Code Repository: \url{https://github.com/Tencent-BAC/YunqueAgent}}}
\end{abstract}

\section{Introduction}
The pursuit of Artificial General Intelligence (AGI) has traditionally prioritized scaling Large Language Models (LLMs) to internalize vast repositories of passive knowledge. However, a critical limitation remains: while these models demonstrate exceptional proficiency in information recall and text generation, they lack the intrinsic agency required to actively discover, verify, and synthesize insights from dynamic, real-world environments. This constraint has necessitated a paradigm shift from static conversational models to autonomous agentic systems capable of grounded problem-solving\cite{yao2023react, shinn2023reflexion, schick2023toolformer}. Central to this evolution is Deep Research—a capability that empowers agents to orchestrate multi-step reasoning, utilize external tools such as web browsers, and navigate open-ended information landscapes autonomously\cite{openai2025deep, geminiteam2025gemini, kimi2025kimiresearcher, hu2025stepdeepresearch, team2025tongyi}. By bridging the gap between pre-trained parameters and active inquiry, Deep Research agents represent a fundamental advancement toward systems that not only emulate human research workflows but substantially augment intellectual productivity.

Despite this promise, realizing the full potential of Deep Research presents significant challenges that prevalent monolithic agent architectures fail to address adequately: 
\textbf{(1) Cognitive Overload in Long-Horizon Tasks} --- 
Long-horizon tasks require maintaining contextual continuity over hundreds of interaction steps. However, in prevailing ReAct-based agents\cite{yao2023react, li2025websailor}, the accumulation of raw execution logs often dilutes the original user intent and degrades reasoning performance.
\textbf{(2) Systemic Fragility and Cascading Failures} --- 
Existing systems often lack resilience. In the absence of robust error detection and recovery mechanisms, minor errors can trigger cascading failures, frequently trapping the entire system in recursive, suboptimal loops\cite{cemri2025why, shen2026metacognitive}.
\textbf{(3) Lack of Modular Extensibility} --- 
The rigid architecture of many contemporary agents inhibits the flexibility required for diverse real-world applications. As research tasks grow in complexity, the inability to seamlessly integrate specialized tools or domain-specific sub-agents limits system composability and adaptability in evolving information environments\cite{luo2025large}.

To address these challenges, we introduce \textbf{Yunque DeepResearch}, a hierarchical agentic framework explicitly engineered to navigate the complexities of deep research.
Our architecture fundamentally restructures the research workflow through a decoupled design that fosters inherent modularity and extensibility, integrating effective sub-agent orchestration and long-horizon context management to transform the typically fragile linear chain into a robust and stable operation. Our system is built upon the following core design principles:

\begin{itemize}
    \item \textbf{Effective Orchestration System.} We implement a centralized orchestration framework anchored by a Main Agent that serves as the strategic core. Utilizing a flexible dispatch mechanism, the planner dynamically routes tasks to the most appropriate resource within the Atomic Capability Pool: it directly invokes basic tools for low-latency atomic operations while delegating complex, long-horizon objectives to specialized sub-agents.
    
    \item \textbf{Dynamic Context Management.} We propose a sub-goal-driven memory mechanism to resolve the tension between context length and information density. By treating sub-goals as the fundamental unit of trajectory segmentation, our system dynamically partitions the research process: completed sub-goals are folded into concise structured summaries to maintain global planning awareness, while the active sub-goal retains fine-grained ReAct traces for precise execution. This hybrid approach transforms linear history into structured semantic milestones.
    
    \item \textbf{Modularity and Extensibility.} We ensure adaptability through a modular ``Atomic Capability Pool'' that separates strategic planning from action execution. By standardizing basic tools and specialized sub-agents as functional units, our architecture attains high composability. This separation creates an extensible ecosystem where new capabilities—ranging from atomic utility functions to expert-level solvers—can be dynamically registered and deployed, ensuring the framework remains resilient to evolving requirements.
    
    \item \textbf{Stability and Robustness.} We incorporate a dedicated Supervisor module to ensure system stability and mitigate the fragility often seen in long-horizon tasks. Unlike rigid reflection schedules, this mechanism performs active anomaly detection on the agent's trajectory. Upon identifying failures, it triggers a self-correction protocol, explicitly prunes invalid context to prevent memory pollution, guiding the agent to autonomously recover and synthesize a viable alternative response.
\end{itemize}

\section{Related Work}
\label{sec:headings}





\paragraph{DeepResearch Agent} 
Recent advancements have witnessed the rapid emergence of Deep Research Agents—notably OpenAI’s Deep Research\cite{openai_deepresearch} and Gemini Deep Research\cite{gemini_deepresearch}—which synergize dynamic reasoning, adaptive planning, and multi-round evidence retrieval to tackle sophisticated open-domain challenges. Within the current research landscape, these systems are objectively categorized into single-agent and multi-agent paradigms based on their architectural orchestration. Conventional single-agent models \cite{openai_deepresearch, gemini_deepresearch, team2025tongyi} typically employ monolithic control architectures to manage tool orchestration. However, such centralized decision-making often suffers from "reliability bottlenecks" and cognitive saturation, particularly when navigating expansive context windows or filtering high-entropy web data. This centralization limits the agent's ability to maintain long-horizon reasoning consistency, leading to performance degradation in complex, multi-step research trajectories. Conversely, multi-agent frameworks \cite{minimax_agent, zhang2025agentorchestra, yu2025aworldorchestratingtrainingrecipe, hu2025owl, oagents} address these limitations by decomposing monolithic objectives into granular, manageable sub-tasks. By leveraging a collective of autonomous, heterogeneous agents—such as dedicated planners, retrievers, and critics—these systems facilitate modular collaboration and explicit error-correction loops. This transition from individual execution to collective orchestration enables superior handling of complex dependencies, effectively mitigating the error propagation typically observed in single-agent configurations.
\paragraph{Working Memory Management} 
Effective management of working memory is crucial to prevent agents from falling into strategic redundancy or infinite loops. Early approaches, such as ReSum\cite{wu2025resum}, utilize specialized summarization models to distill history upon reaching context limits. Subsequent works, including MeM1\cite{2025mem1learningsynergizememory} and Memory-as-Action\cite{zhang_memory_2025}, treat memory management as an explicit reasoning action, enabling agents to autonomously summarize or edit their internal states. To handle ultra-long sequences, MemAgent\cite{yu2025memagent} selectively stores task-relevant information, while AgentFold\cite{ye_agentfold_2025} adaptively determines compression granules to balance information density. Despite these advances, existing methods primarily focus on content compression rather than structured synthesis, often failing to maximize the effective information density of the agent's historical trajectory.


\paragraph{Specialized Agent} 
Beyond general-purpose research, specialized agents have been developed for domain-specific tasks. GUI Agents (e.g., CogAgent\cite{hong2023cogagent}, Claude 3.5 Sonnet\cite{cluade_3.5_sonnet}) autonomously interact with digital systems by perceiving and manipulating visual elements. Similarly, Data Analysis Agents (e.g., OpenAI’s Advanced Data Analysis\cite{openai_data_analysis}, Data Interpreter\cite{hong2025data}) automate the end-to-end pipeline of data cleaning, processing, and reporting. Notably, these specialized agents are increasingly integrated as functional modules within larger multi-agent frameworks (e.g., OpenAgents\cite{xie2023openagents}, AgentOrchestra\cite{zhang2025agentorchestra}) to execute sophisticated data-centric sub-goals.


\section{Framework}
\label{sec:framework}







\begin{figure}
  \centering
  \includegraphics[width=0.8\textwidth]{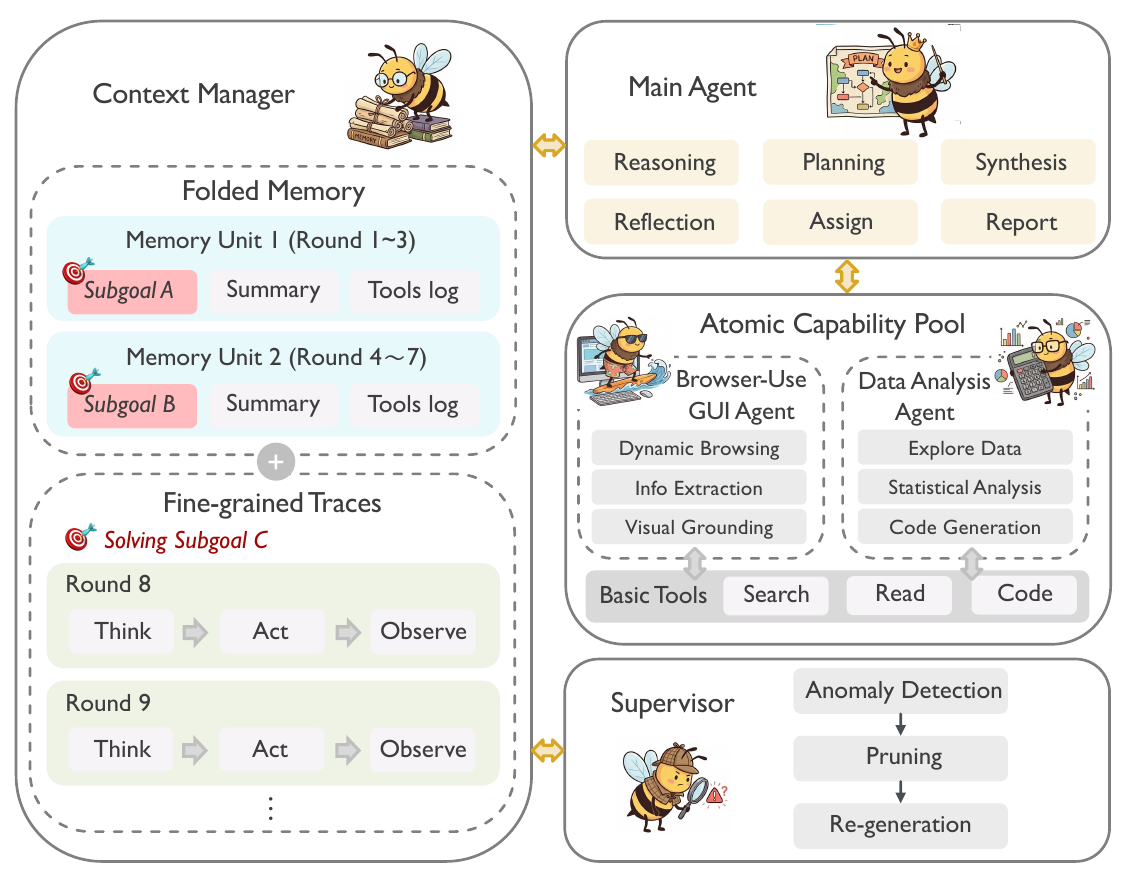}
  \caption{Overview of Yunque DeepResearch.}
  \label{fig:framework}
\end{figure}

As illustrated in Figure \ref{fig:framework}, we propose Yunque DeepResearch, a hierarchical architecture designed to explicitly decouple atomic capabilities from the central reasoning process. The framework comprises four collaborative modules: \textit{(i) Main Agent, (ii) Context Manager, (iii) Atomic Capability Pool, (iv) Supervisor}.

\begin{itemize}
    \item \textbf{Main Agent}: At the core of the system, this module acts as the central executive, responsible for high-level intent recognition, dynamic planning, global orchestration, and result synthesis.
    
    \item \textbf{Context Manager}: To support navigating long-horizon tasks, this module establishes a dual-level memory structure that balances immediate operational precision with long-term strategic context.
    
    \item \textbf{Atomic Capability Pool}: Complementing the reasoning core, this pool hosts a diverse set of specialized sub-agents (e.g., Browser-Use GUI and Data Analysis agents) and fundamental tools, enabling the execution of specific atomic actions.
    
    \item \textbf{Supervisor}: This module ensures system robustness by monitoring the execution trajectory, performing error correction, and preventing cascading failures during multi-turn interactions.
\end{itemize}

\subsection{Main Agent}

The Main Agent focuses on decomposing high-level user instructions into manageable sub-tasks through intent analysis and logical structuring. Adopting an interleaved reasoning paradigm, the agent actively interprets feedback from each step to dynamically refine its subsequent plans, persisting reasoning context across tool invocations and sub-agent handoffs to maintain global coherence.

To execute these plans, the Main Agent employs an adaptive routing mechanism that synergizes lightweight tools with specialized sub-agents. Instead of adhering to a rigid division of labor, the agent dynamically orchestrates these complementary resources based on the evolving context. This design allows for a fluid interplay between direct tool invocations and sub-agent delegation, ensuring that execution complexity is effectively encapsulated while maintaining the flexibility to address diverse problem constraints.



\subsection{Memory}
The primary challenge in historical information compression is optimizing the granularity of salient information—minimizing loss from over-compression while maintaining a high-density information structure. To address this, we propose a working-memory management method comprising two core components: \textbf{Structured Memory Generation}, which handles the construction and dynamic updating of structured memory states, and \textbf{Dynamic Context Management}, which is responsible for dynamically and directly adjusting the context based on structured memory, providing the LLM with the most valuable and information-dense input.



\subsubsection{Structured Memory Generation}
We propose a structured memory architecture that prioritizes high-value history information along three dimensions: \textbf{sub-goal} (planning objectives), \textbf{tools-log} (tool invocation strategies), and \textbf{summary} (summary of information highly relevant to the query). Unlike linear history, our mechanism achieves high-dimensional abstraction by partitioning the agent's trajectory into semantically cohesive units based on sub-goals.

\paragraph{Memory Unit Structure}
Specifically, each memory unit is defined as a 4-tuple:
\begin{equation} 
    m_i = (\mathcal{R}_i, g_i, \mathcal{T}i, s_i) 
\end{equation} 
where: 
\begin{itemize}
   \item $\mathcal{R}_i$ is the list of global round index contributing to the sub-goal.
   \item $g_i$ is the semantic description of the current sub-goal, providing short-term guidance.
   \item $\mathcal{T}_i$ is the list of persistent tool-use log, consisting of tool names, parameters, and execution results.
   \item $s_i$ is an incremental summary of key information extracted during task execution.
\end{itemize}
The complete memory list $\mathcal{M} = [m_1, m_2, \dots, m_n]$ represents a macro-level planning pathway that drives the agent toward task completion.

\paragraph{Dynamic Folding and Adding Mechanism} 
An end-to-end memory model $\mathcal{F}_{\text{mem}}$ is introduced to manage transitions between sub-goals. At each interaction round $t$, $\mathcal{F}_{\text{mem}}$ processes the latest main-agent's response $a_t$, action resulting observation $o_t$ and the latest memory unit $m_n$($t>1$) to produce a binary indicator $\delta_{\text{fold}}$ and an updated unit $m_{\text{out}}$:
\begin{equation} 
(m_{\text{out}}, \delta_{\text{fold}}) = \mathcal{F}_{\text{mem}}(a_t, o_t, m_n)
\end{equation}

When $\delta_{\text{fold}} = 1$, it indicates the sub-goal of the current round is consistent with $g_n$. In this case, $m_{\text{out}}$ as the latest updated memory unit, will directly replace the last element of the existing memory list:
\begin{equation} 
\mathcal{M} \leftarrow \mathcal{M}_{1:n-1} \oplus [m_{\text{out}}]
\end{equation} 
where $m_{\text{out}}$ incorporates the latest round index, the latest updated tool log $\tau_t$ and the latest updated summary $s_n$.

When $\delta_{\text{fold}} = 0$, it indicates that the sub-goal of the current round differs from the existing latest memory unit. In this case, $m_{\text{out}}$ as a newly constructed memory unit will be appended to the memory list:
\begin{equation} 
\mathcal{M} \leftarrow \mathcal{M} \oplus [m_{\text{out}}]
\end{equation} 
where $m_{\text{out}}$ is constructed using the new sub-goal $g_t$, initial tool log $\tau_t$ and initial summary $s_n$.

This mechanism effectively compresses execution trajectories by aggregating multi-round interactions into single intentional units, significantly reducing context redundancy while preserving a high-fidelity history of the decision-making process.

\subsubsection{Dynamic Context Management}
Leveraging the structured memory $\mathcal{M}$, we propose a dynamic context management strategy to mitigate the linear context growth typical of the ReAct paradigm. By adaptively switching between fine-grained execution traces and compressed memory units, our approach ensures information integrity while optimizing efficiency.

\paragraph{Adaptive Context Construction}
Let $Q$ be the user query and $|\mathcal{R}_n|$ be the  count of the current memory unit. The context $\mathcal{C}_t$ can be constructed in the following two formats alternatively:


\begin{equation} 
\mathcal{C}_{t} =
\begin{cases} 
\mathcal{C}_{t-1} \oplus [r_t, o_t] & \text{if } |\mathcal{R}_{n}| > 1 \\[6pt]
(Q, \mathcal{M}_{1:n-1}) \oplus [r_t, o_t] & \text{if } |\mathcal{R}_{n}| = 1 
\end{cases} 
\end{equation}

When $|\mathcal{R}_n| > 1$, the agent is actively pursuing the current sub-goal $g_n$. We retain the incremental ReAct format, appending the latest response $r_t$ and observation $o_t$ to $\mathcal{C}_{t-1}$. This preserves the fine-grained execution trace necessary for tracking local dependencies within a sub-goal.

When $|\mathcal{R}_n| = 1$, it signals that the agent is starting a new sub-goal. To prevent redundant contextual information, we perform a compression reset, replacing historical round-by-round in model ReAct with serialized folded memory of all completed sub-goals $\mathcal{M}_{1:n-1}$.

\paragraph{Efficiency and Complexity} Our memory management shifts the context complexity from $\mathcal{O}(t)$ total rounds to $\mathcal{O}(n)$ sub-goals. By triggering compression only upon sub-goal completion ($|\mathcal{R}_n|=1$), the mechanism maintains a balance between high-level task awareness and low-level operational detail, ensuring that cross-sub-goal information is retained without incurring the redundancy of linear history.

\subsection{Atomic Capability Pool}

To construct a robust generalist assistant capable of navigating complex real-world workflows, we designed the atomic capability pool as a hierarchical system comprising two layers: \textit{(i) Specialized Sub-agents}, which encapsulate high-level reasoning and multi-step planning for specific domains (GUI interaction and Data Analysis); and \textit{(ii) Basic Tools} (detailed in Appendix C), which provide fundamental primitives for information retrieval and execution.

\subsubsection{Browser-Use GUI Agent}
To better align with our actual system implementation, we refer to this specialized sub-agent as the Browser-Use GUI Agent. Its interaction environment is restricted to a web browser, and it completes web-based subtasks through a closed loop of ``observe--act--feedback'', including searching, clicking, scrolling, typing, and content extraction on dynamic webpages. Unlike text-only fetching pipelines, this agent explicitly models \textit{interactive elements} and \textit{multi-tab states}, enabling robust information acquisition in scenarios that require real interaction (e.g., pagination, collapsible sections, login pop-ups, and lazy-loaded content).

\paragraph{Problem Formulation (POMDP).}
We model the interaction of the Browser-Use GUI Agent as a partially observable sequential decision process (POMDP), represented by the tuple
\begin{equation}
\mathcal{E}=\langle \mathcal{U},\mathcal{A},\mathcal{S},\mathcal{O},T \rangle ,
\end{equation}
where $\mathcal{U}$ denotes the task space (including user instructions and sub-goals), $\mathcal{A}$ the action space, $\mathcal{S}$ the underlying environment state space (not fully observable to the agent), $\mathcal{O}$ the observation space, and $T:\mathcal{S}\times \mathcal{A}\rightarrow \mathcal{P}(\mathcal{S})$ the state transition function.

\paragraph{State and Observation.} In the browser setting, we define the true state $s_t\in\mathcal{S}$ as the complete information of the current browser environment. It includes not only structured web information (the full DOM tree, page contents of all tabs, the URL and navigation history, scroll and focus states, dynamic scripts and network request states, etc.), but also complete visual information (pixel-level rendering, layout and styles, and content outside the current window, including collapsed regions and visual content that becomes visible only after scrolling). This true state is not fully observable to the agent.

The observation $o_t\in\mathcal{O}$ obtained by the agent at step $t$ consists of three parts:
\begin{equation}
o_t = (c_t, b_t, x_t),
\end{equation}
where $c_t$ is the textual context required to solve the task (e.g., the user query, the current sub-goal, key historical findings, and the previous action with its returned results), which contains no image information; $b_t$ is a structured browser-state extracted from the currently visible window (i.e., the viewport), including the URL, title, tab list, scroll information, and interactive element indices with their corresponding DOM snippets limited to the current viewport; and $x_t$ is a screenshot of the currently opened page (full-page rendering). Since $b_t$ only covers the visible viewport (content outside the window cannot be directly ``seen''), the process is inherently partially observable.

Importantly, in our implementation, the screenshot $x_t$ is provided as ephemeral multi-modal input only for the current step, and is not serialized into the textual context history, preventing context explosion in long-horizon interactions.

Parameterized Discrete Actions (Tool-call). We represent each browser operation as a parameterized action
\begin{equation}
a_t = (\alpha_t,\theta_t)\in\mathcal{A},
\end{equation}
where $\alpha_t$ specifies the action type (e.g., \texttt{web\_search}, \texttt{pdf\_to\_markdown}, \texttt{go\_to\_url}, \texttt{click\_element}, \texttt{input\_text}, \texttt{scroll\_down/up}, \texttt{extract\_content}, \texttt{open/switch/close\_tab}, \texttt{terminate}), and $\theta_t$ contains structured parameters (e.g., URL, element index, input text, scroll pixels). In particular, for \texttt{pdf\_to\_markdown}, we adopt a paging mechanism to read long PDF documents incrementally: the converted Markdown is retrieved in pages (using an offset and a maximum character budget per page), rather than injected in full, which prevents context explosion in long-horizon interactions. In implementation, we enforce a ``one tool per turn'' constraint: at most one tool-call is executed per step, decomposing complex web interaction into a multi-round sequence of atomic decisions.

\paragraph{Policy.} Due to partial observability, the policy $\pi$ selects the next action based on the interaction history. Let
\begin{equation}
h_t = (u, o_1,a_1,\dots,o_t)
\end{equation}
denote the interaction history up to step $t$ (including the task $u\in\mathcal{U}$, the observation sequence, and executed actions). The policy is defined as
\begin{equation}
\pi(a_t \mid h_t): \Delta(\mathcal{H}) \rightarrow \mathcal{A},
\end{equation}
where $\mathcal{H}$ is the history space and $\Delta(\mathcal{H})$ denotes the probability simplex over it. The environment transitions according to $T$ as $s_{t+1}\sim T(s_t,a_t)$ and produces the next observation $o_{t+1}$; the interaction terminates when \texttt{terminate} is selected or when a maximum step budget is reached.

\subsubsection{Data Analysis Agent}
\begin{figure}
  \centering
  \includegraphics[width=0.9\textwidth]{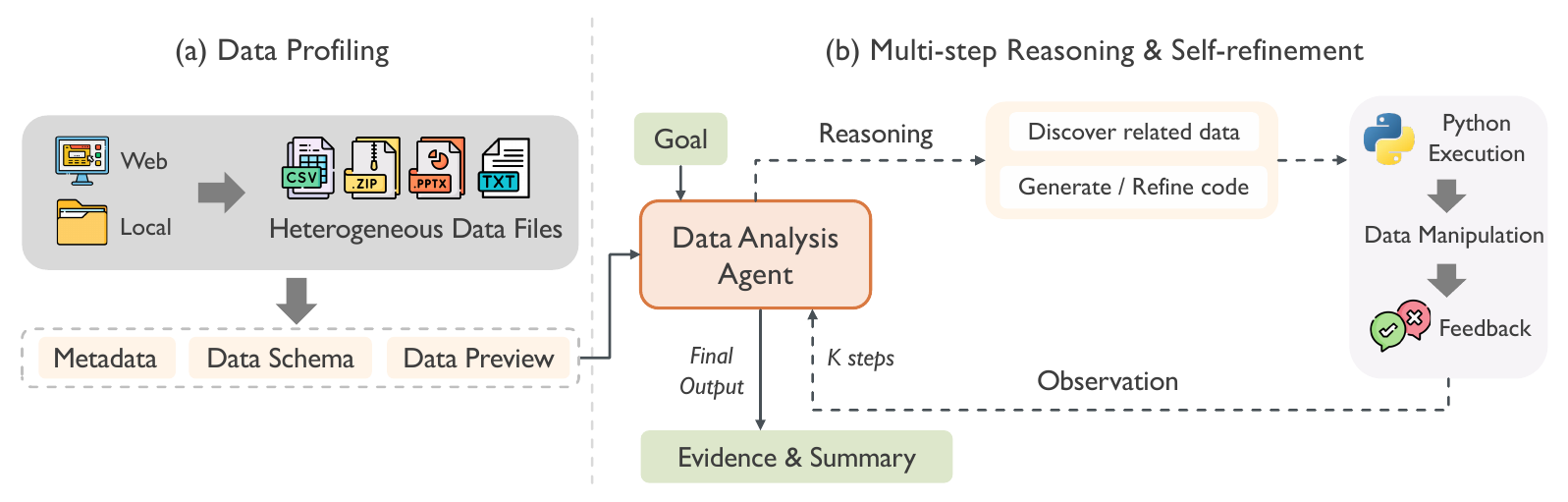}
  \caption{Illustration of Data Analysis Agent.}
  \label{fig:da-agent}
\end{figure}

The Data Analysis Agent is designed to automate complex data tasks by leveraging a multi-step reasoning process and generating executable code to perform intricate data processing, refine execution outputs, and generate evidence from heterogeneous sources. As illustrated in Figure \ref{fig:da-agent}, the workflow is structured into two primary phases: (a) Data Profiling, which yields comprehensive descriptions and previews of the input data, and (b) an iterative loop of multi-step reasoning and self-refinement to derive answers.

\paragraph{Data Profiling.} 
The initial phase focuses on the comprehensive profiling of heterogeneous data files originating from web or local storage. The agent is engineered to handle a wide array of formats, including structured data, unstructured text, and compressed archives. Upon receiving files, the agent employs a routing mechanism to invoke appropriate tools—such as Pandas for tabular data or specialized parsers for documents—to extract essential information. This process yields a structured description comprising three key components: \textit{(i) Metadata} (e.g., file name, size, type), \textit{(ii) Data Schema} (e.g., dimensions, column names), and \textit{(iii) Data Preview} (e.g., content snippets or full content, statistical summaries). This profiling step transforms raw files into a standardized context ready for analysis.

\paragraph{Multi-step Reasoning and Self-refinement.} 
Building upon the meticulously compiled data profiles, the agent enters an iterative loop of reasoning and refinement driven by the user's goal. This phase integrates the following sequential processes: 
\textit{(i) Reasoning and Code Generation:} Based on the profiled context and current goal, the agent reasons to confirm file relevance, discover related data sections, and generate executable Python scripts to manipulate the data. 
\textit{(ii) Execution and Feedback:} The generated code is executed within a secure sandbox, where the system captures the execution output or error message as feedback. 
\textit{(iii) Observation and Self-refinement:} The agent assesses this feedback; if an error occurs, it triggers self-refinement to revise the code, whereas a successful observation informs the next reasoning step. 
This cycle continues until the objective is met, at which point the agent synthesizes the findings to produce the final output, which presents the supporting evidence and a final, coherent summary.

\subsection{Adaptive Interrupt and Self-Correction Mechanism}

In complex, long-horizon reasoning tasks, agents often succumb to cognitive inertia, where they persist in invalid behaviors despite execution failures. Common symptoms include syntactic errors (e.g., malformed tool calls) and semantic stagnation (e.g., repetitive outputs or recursive loops), which prevent the agent from progressing toward the solution.

To mitigate this, we introduce a Supervisor module equipped with an \textit{Adaptive Interrupt and Self-Correction Mechanism} into the framework. Unlike rigid reflection schedules that might interfere with valid reasoning chains, the Supervisor continuously evaluates the agent's trajectory for anomalies. Upon detecting signals of failure or stagnation, it triggers a preemptive interrupt, forcibly transitioning the agent from an automated \textit{Acting Mode} to a high-level \textit{Reflective Mode}. In this state, the system executes a three-stage recovery protocol:

\begin{itemize} 
\item \textbf{Anomaly Diagnosis:} 
The agent is prompted to critically analyze the execution history to pinpoint the root cause of failure.
\item \textbf{Trajectory Pruning:} 
To prevent context pollution, the system explicitly prunes the recent invalid interaction traces from the context window. This ensures that the agent's future decisions are not biased by the failed attempts.
\item \textbf{Re-generation:}
The agent synthesizes an alternative output—such as a revised plan or a corrected conclusion—effectively breaking the local loop and guiding the workflow back to a valid solution path.
\end{itemize}

\section{Experiments}
\paragraph{Experimental Setup.}
In all experiments, we maintained consistent hyperparameters across comparable settings, setting the temperature to $1.0$ and top-p to $0.95$. The time allowance for each question is limited to $1.5$ hours, and the maximum number of tool calls is capped at $75$. All results were obtained in January 2026. Unless otherwise specified, all models employed within the framework are based on Gemini-3-pro to ensure consistent model capability across ablation studies. The maximum number of steps for specialized sub-agents is set to $10$.

\paragraph{Benchmarks.}
We adhere to the official evaluation protocols for each benchmark. Our evaluation encompasses: (1) GAIA\cite{mialon2023gaia}; (2) BrowseComp\cite{wei2025browsecomp}; (3) BrowseComp-ZH\cite{zhou2025browsecompzh}; and (4) Humanity's Last Exam\cite{phan2025humanitys}. Detailed evaluation settings are provided in Appendix B.

\paragraph{Evaluation Metrics.}
The primary metric is \textit{Pass@N}, defined as the probability that at least one correct solution appears among $N$ independent executions. Unless otherwise specified, we report the \textit{Pass@1} score, reflecting the model's ability to produce a correct answer in a single attempt.

\paragraph{Baselines.}
Our comparative analysis involves baselines categorized into three groups: (1) \textit{General LLMs utilizing ReAct and basic tools}, including GLM-4.7\cite{glm-4.5team2025glm45}, Kimi-K2 thinking\cite{kimiteam2025kimi}, DeepSeek-V3.2\cite{deepseek-ai2025deepseekv32}, MiniMax M2\cite{minimax-aiminimax}, Claude Sonnet 4.5, GPT-5 High, OpenAI-o3 and Gemini 3 Pro; (2) \textit{Open-source Agent Frameworks} such as DeepAgent\cite{li2025deepagent}, OAgent\cite{zhu2025oagents}, MiroFlow\cite{miromindaimiroflow}, AgentFold\cite{ye2025agentfold}, Tongyi DeepResearch\cite{team2025tongyi}, and WebResearcher\cite{qiao2025webresearcher}; and (3) \textit{Closed-source Agent Frameworks} including Kimi-Researcher\cite{kimi2025kimiresearcher}, Gemini Deep Research\cite{geminiteam2025gemini}, and OpenAI Deep Research\cite{openai2025deep}. These selections cover a wide range of existing capabilities, establishing a solid foundation for benchmarking the performance of the Yunque DeepResearch framework.

\subsection{Main Results}

Table~\ref{tab:main_results} presents a comprehensive comparison of Yunque DeepResearch against state-of-the-art baselines across four demanding benchmarks. Our framework demonstrates superior performance, achieving the highest scores on BrowseComp (62.5), BrowseComp-ZH (75.9), and Humanity's Last Exam (51.7), while securing a competitive second place on GAIA (78.6).

Comparative analysis reveals two key insights. First, Yunque DeepResearch significantly enhances the capabilities of base models. For instance, it boosts Gemini 3 Pro's performance by +10.0 on BrowseComp and +4.8 on GAIA compared to the standard ReAct baseline, demonstrating that our agentic design effectively unlocks the underlying LLM's potential. Second, our framework exhibits exceptional competitiveness against existing agentic systems; it consistently outperforms both open-source and closed-source baselines, validating its effectiveness in open-ended research.

\begin{table}[htbp]
\centering
\caption{Performance comparison on various agentic benchmarks. \textbf{Bold} and \underline{underlined} indicate the best and second-best results, respectively. Results marked with $^*$ are sourced from official technical reports.}
\label{tab:main_results}
\resizebox{0.75\textwidth}{!}{
\begin{tabular}{lccccc}
\toprule
\textbf{Benchmarks} & \textbf{Browse} & \textbf{Browse} & \textbf{GAIA} & \textbf{Humanity's} & \textbf{Model} \\
& \textbf{Comp} & \textbf{Comp-ZH} &  & \textbf{Last Exam} & \textbf{Family}\\
\midrule
\multicolumn{6}{l}{\textit{General LLMs + ReAct + Basic Tools}} \\
\midrule
GLM 4.7 & 52.0$^*$ & 66.6$^*$ & -- & 42.8$^*$ & --\\
Kimi K2 Thinking & \underline{60.2}$^*$ & 62.3$^*$ & 69.9 & 44.9$^*$ & --\\
DeepSeek-V3.2 & 51.4$^*$ & 65.0$^*$ & 70.9 & 40.8$^*$ & --\\
MiniMax-M2 & 44.0$^*$ & -- & 75.7$^*$ & -- & --\\
Claude-4.5-Sonnet & 24.1$^*$ & 42.4$^*$ & 71.2$^*$ & 32.0$^*$ & --\\
GPT-5 High & 54.9$^*$ & 63.0$^*$ & 76.4$^*$ & 35.2$^*$ & --\\
OpenAI-o3 & 49.7$^*$ & 58.1$^*$ & 70.5$^*$ & 20.2$^*$ & --\\
Gemini 3 Pro & 52.5 & \underline{67.1} & 73.8 & \underline{45.8}$^*$ & --\\
\midrule
\multicolumn{6}{l}{\textit{Open-source Agent Frameworks}} \\
\midrule
DeepAgent & -- & -- & 58.3$^*$ & 20.2$^*$ & QwQ\\
Tongyi DeepResearch & 43.4$^*$ & 46.7$^*$ & 70.9$^*$ & 32.9$^*$ & Qwen\\
AgentFold & 36.2$^*$ & 47.3$^*$ & 67.0$^*$ & -- & Qwen \\
WebResearcher {\scriptsize (w/o TTS)} & 37.3$^*$ & 45.2$^*$ & 72.8$^*$ & 28.8$^*$ & Qwen \\
OAgent & 22.2$^*$ & -- & 66.7$^*$ & 15.4$^*$ & Claude\\
MiroFlow & 33.2$^*$ & 47.1$^*$ & \textbf{82.4}$^*$ & 27.2$^*$ & GPT\\
\midrule
\multicolumn{6}{l}{\textit{Close-source Agent Frameworks}} \\
\midrule
Gemini DeepResearch & -- & -- & -- & 26.9$^*$ & Gemini\\
Kimi Researcher & -- & -- & -- & 26.9$^*$ & Kimi\\
OpenAI DeepResearch & 51.5$^*$ & 42.9$^*$ & 67.4$^*$ & 26.6$^*$ & GPT\\
\midrule
\rowcolor{yellow!10} Yunque DeepResearch & \textbf{62.5} & \textbf{75.9} & \underline{78.6} & \textbf{51.7} & Gemini\\
\bottomrule
\end{tabular}
}
\end{table}

\subsection{Detailed Analysis}

\paragraph{Generalizability Across Backbones.}

To verify that the performance gains stem from our framework design rather than solely the capability of the underlying model, we evaluated Yunque DeepResearch using different backbone models: DeepSeek-V3.2, Kimi K2 Thinking, and Gemini 3 Pro. As shown in Table~\ref{tab:backbone}, our framework consistently enhances performance across diverse model families. Specifically, DeepSeek-V3.2 and Gemini 3 Pro exhibit robust improvements on all benchmarks, achieving peak gains of +10.1 (on BrowseComp-ZH) and +10.0 (on BrowseComp), respectively. While Kimi K2 Thinking shows a marginal fluctuation on BrowseComp-ZH, it secures a substantial +4.9 improvement on GAIA. We note that we selected these representative backbones based on their robust instruction-following capabilities, as precise tool invocation is a prerequisite for our framework that not all models currently satisfy. Collectively, these results confirm that Yunque DeepResearch is a model-agnostic solution that universally enhances problem-solving abilities.

\begin{table}[!h]
\centering
\setstretch{1.1}
\caption{Evaluation of our framework applied to different backbone models. Results marked with $^*$ are sourced from official technical reports.}\label{tab:backbone}
\begin{tabular}{l|c|c|c|c|c}
\toprule
\textbf{Model} & \textbf{Mode}
    & \textbf{GAIA} & \textbf{BrowseComp-ZH} & \textbf{BrowseComp} & \textbf{HLE} \\
\midrule
\multirow{3}{*}{DeepSeek-V3.2}
& ReAct & 70.9 & 65.0$^*$ & 51.4$^*$ & 40.8$^*$ \\
& Ours & 76.7 & 75.1 & 59.5 & 46.6  \\
& Gap   & \uag{5.8} & \uag{10.1} & \uag{8.1} & \uag{5.8}  \\
\midrule
\multirow{3}{*}{Kimi K2 Thinking}
& ReAct & 69.9 & 62.3$^*$ & -- & 44.9$^*$  \\
& Ours & 74.8 & 61.9 & -- & -- \\
& Gap   & \uag{4.9} & \dab{0.4} & -- & --  \\
\midrule
\multirow{3}{*}{Gemini 3 Pro}
& ReAct & 73.8 & 67.1 & 52.5 & 45.8$^*$ \\
& Ours & 78.6 & 75.9 & 62.5 & 51.7 \\
& Gap   & \uag{4.8} & \uag{8.8} & \uag{10.0} & \uag{5.9} \\
\bottomrule
\end{tabular}
\end{table}

\paragraph{Efficacy of Modular Design.}
We conducted an ablation study (Table~\ref{tab:ablation}) to assess the contribution of each component within Yunque DeepResearch. The results validate the necessity of our modular design:

\begin{itemize}
    \item \textbf{Impact of Memory.} Removing the memory module results in the sharpest performance drop on browsing tasks ($-10.4$ on BrowseComp and $-7.4$ on BrowseComp-ZH). Since these benchmarks necessitate long-horizon information seeking and multi-hop reasoning, this decline underscores the critical role of our memory mechanism in mitigating the high noise ratio inherent in web browsing. While standard ReAct paradigms often suffer from context overflow and ``lost-in-the-middle'' phenomena—where valid information is submerged by invalid attempts—our memory management preserves information density by partitioning the agent's trajectory into semantically cohesive units via sub-goals.
    \item \textbf{Impact of Supervisor.} The Supervisor acts as a critical safeguard for complex task execution, with its removal precipitating significant performance declines across GAIA ($-8.7$), BrowseComp-ZH ($-10.5$), BrowseComp ($-4.4$), and HLE ($-1.2$). This module is essential for mitigating execution fragility in rigorous environments. Instead of allowing the agent to spiral into failure loops caused by malformed tool calls or ineffective actions, the Supervisor enforces a high-level orchestration layer that actively monitors for failure signals. By intercepting these errors and clearing invalid interaction traces from the context window, it prevents ``error accumulation'' and ensures the agent's reasoning process remains unpolluted. This mechanism guarantees that subsequent attempts are not biased by previous failures, thereby maintaining a clean and valid solution path during long-horizon reasoning.
    \item \textbf{Impact of Specialized Agents.} We prioritize the GAIA benchmark for this analysis, as its heterogeneous task composition—spanning file processing, data analysis, and open-world browsing—imposes a unique demand for domain-specific competencies. The ablation of the Browser-Use GUI Agent and Data Analysis Agent resulted in notable performance regressions ($-6.8$ and $-2.9$, respectively). This degradation corroborates the hypothesis that robust ``general'' assistance emerges from the orchestration of specialized atomic capabilities. These specialized modules provide the precise, low-level execution skills that monolithic models often lack when confronting the multimodal diversity of real-world tasks.
\end{itemize}

\begin{table}[!h]
\centering
\setstretch{1.1}
\caption{Ablation study assessing the impact of individual components.}\label{tab:ablation}
\begin{tabular}{lcccc}
\toprule
\textbf{Method} & 
    \textbf{GAIA} & \textbf{BrowseComp} & \textbf{BrowseComp-ZH} & \textbf{HLE} \\
\midrule
\rowcolor{yellow!10}Yunque DeepResearch
 & \textbf{78.6} & \textbf{62.5} & \textbf{75.9} & \textbf{51.7} \\
\midrule
$\quad$w/o Memory
 & 77.7 & 52.1 & 68.5 & 51.7  \\
 & \dab{0.9} & \dab{10.4} & \dab{7.4} & -- \\
\midrule
$\quad$w/o Supervisor
 & 69.9 & 58.1 & 65.4 & 50.5 \\
 & \dab{8.7} & \dab{4.4} & \dab{10.5} & \dab{1.2} \\
\midrule
$\quad$w/o Browser-Use GUI Agent
 & 71.8 & 61.7 & -- & -- \\
 & \dab{6.8} & \dab{0.8} & -- & -- \\
\midrule
$\quad$w/o Data Analysis Agent
 & 75.7 & -- & -- & 51.7 \\
 & \dab{2.9} & -- & -- & -- \\
\bottomrule
\end{tabular}
\end{table}

\section{Limitations and Future Work}

While our framework demonstrates promising results, we acknowledge several limitations that outline directions for future research. First, regarding the evaluation scope, although our main experiments encompass a diverse set of benchmarks, the ablation studies for specialized sub-agents are primarily anchored on GAIA. While GAIA serves as a robust testbed for generalist capabilities, it may not fully capture the nuances of specialized domains. Second, we have not yet conducted a systematic analysis of token consumption and inference latency. Although we explicitly designed sub-agents as lightweight workflows—characterized by concise instructions, finite execution horizons, and reduced context dependency—our empirical observations suggest that total execution time remains heavily dependent on the reasoning capability of the underlying foundation model.

In future work, we aim to extend the assessment of our specialized agents to include domain-specific benchmarks, such as DSBench\cite{jing2025dsbench} for comprehensive data analysis assessment and OSWorld\cite{xie2024osworld} for complex, interactive environment navigation, thereby rigorously testing their performance boundaries. Furthermore, we plan to explore post-training strategies for atomic capabilities to develop smaller and more efficient modules, thereby reducing latency and computational costs.

\section{Conclusion}
We presented Yunque DeepResearch, a hierarchical, modular, and robust multi-agent framework designed to address the critical challenges of cognitive overload in long-horizon tasks, systemic fragility in autonomous execution, and the lack of modular extensibility in rigid architectures. Extensive empirical evaluations demonstrate that Yunque DeepResearch achieves state-of-the-art results on challenging benchmarks, including GAIA, BrowseComp, BrowseComp-ZH, and Humanity’s Last Exam. We hope that our open-source contributions will provide a solid foundation for future research on collaborative agentic systems.

\newpage
\section*{\Large Appendix}
\section*{A $\;$ Case Study}
In this section, we select one question from the BrowseComp dataset as an example of a long-horizon task to illustrate our memory generation process in detail. Additionally, we select a challenging information retrieval case from the GAIA dataset as an example, showcasing our agent's problem-solving workflow to illustrate the superiority of proposed hierarchical framework.
\paragraph{Example 1: Memory}

The following question selected is a typical cross-domain and multi-step reasoning challenge. Its reasoning path involves multi-level logical progression, where the failure to verify any minor detail can lead to the systemic collapse of the entire reasoning chain. Our dynamic memory management mechanism achieves a structured evolution of memory storage by capturing the real-time strategies of the main agent. Each memory unit encapsulates key information consisting of 'Goal-Action Process-Outcome'. Based on sub-goals, these memory units form a continuous reasoning chain that reflects the agent's systematic 'Investigation-Reasoning-Verification' workflow.





\begin{tcolorbox}[breakable,title=Case Memory Generation for Example 1 in BrowseComp]
\textbf{Question:} There's a thesis submitted between 2020 and 2023, inclusive, for a Doctor of Philosophy degree at a university established between January 1st, 1980, and December 31, 1990, inclusive. The author dedicated the thesis to their children and the thesis is related to dating apps. In its acknowledgment, the author mentioned about their committed relationship coming to an end and starting a podcast. The author started this podcast with someone they originally met at a film event. What's the name of the podcast?

\vspace{0.2cm}

\noindent\colorbox{gray!70}{%
\parbox{\dimexpr\textwidth-2\fboxsep\relax}{%
\centering
\textbf{[Memory Unit 1]: Round 1}%
}%
}

\vspace{0.2cm}

\textbf{sub-goal:} Investigate Piia Myyry as the potential author and identify the podcast name

\vspace{2mm}
\textbf{tools-log:}
\begin{lstlisting}[language=json]
[{
    "tool": "search",
    "args": {
        "query": ["PhD thesis \"dating apps\" dedication \"children\" 2020..2023", "PhD thesis \"dating apps\" acknowledgment \"podcast\" \"film event\"", "PhD thesis \"dating apps\" acknowledgment \"relationship\" \"podcast\"", "PhD thesis \"dating apps\" dedicated to my children"]
    }",
    "status": "success"
}]
\end{lstlisting}

\textbf{summary:} The agent executed searches for specific phrases in thesis acknowledgments. While the agent's thought process initially focused on Lisa Portolan, the tool output surfaced a strong lead for 'Piia Myyry', whose thesis is associated with the phrases 'met at a film event' and 'started a podcast' in the context of dating apps.

\vspace{0.2cm}

\noindent\colorbox{gray!70}{%
\parbox{\dimexpr\textwidth-2\fboxsep\relax}{%
\centering
\textbf{[Memory Unit 2]: Round 2\textasciitilde3}%
}%
}

\vspace{0.2cm}

\textbf{sub-goal:} Verify Lisa Portolan as the author and identify the specific podcast

\vspace{2mm}
\textbf{tools-log:}
\begin{lstlisting}[language=json]
[{
    "tool": "search",
    "args": {
        "query": ["Lisa Portolan \"Slow Love\" podcast met co-host film event", "Lisa Portolan thesis acknowledgments \"relationship\" \"podcast\"", "Lisa Portolan \"Slow Love\" podcast \"breakup\"", "Lisa Portolan PhD thesis dedication children"]
    },
    "status": "success"
}]
\end{lstlisting}

\textbf{Summary:} Search results confirm she has a PhD from Western Sydney University (established 1989) focusing on dating apps and has a podcast called 'Slow Love'. The next steps focus on finding the full text or specific acknowledgments of her thesis to confirm the personal details and the podcast name.

\vspace{0.2cm}

\noindent\colorbox{gray!70}{%
\parbox{\dimexpr\textwidth-2\fboxsep\relax}{%
\centering
\textbf{[Memory Unit 3]: Round 4\textasciitilde5}%
}%
}

\vspace{0.2cm}

\textbf{sub-goal:} Identify the podcast name by searching for specific phrases from the thesis acknowledgment

\vspace{2mm}
\textbf{tools-log:}
\begin{lstlisting}[language=json]
[
    {
        "tool": "search",
        "args": {
            "query": ["\"started a podcast with whom I first met at a film event\"", "\"my committed relationship came to an end\" podcast thesis", "Lisa Portolan thesis acknowledgments \"film event\"", "Lisa Portolan \"Slow Love\" podcast co-host film event", "PhD thesis dating apps \"dedicated to my children\""]
        },
        "status": "success"
    },
    {
        "tool": "search",
        "args": {
            "query": ["\"started a podcast with whom I first met at a film event\"", "\"my committed relationship came to an end\" thesis", "\"A Lifestyle Gap\" PhD thesis dating apps", "Lisa Portolan thesis \"film event\"", "podcast co-host \"met at a film event\""]
        },
        "status": "success"
    }
]
\end{lstlisting}

\textbf{summary:} The agent confirmed the unique text string: 'my committed relationship came to an end and i started a podcast with whom i first met at a film event'. Piia Myyry and 'A Lifestyle Gap' appear in snippets but may be noise; the agent suspects the author is Lisa Portolan.

\vspace{0.2cm}

\noindent\colorbox{gray!70}{%
\parbox{\dimexpr\textwidth-2\fboxsep\relax}{%
\centering
\textbf{[Memory Unit 4]: Round 6}%
}%
}

\vspace{0.2cm}

\textbf{sub-goal:} Verify the thesis title 'A Lifestyle Gap' and the podcast name by confirming the author's identity and the 'film event' origin story

\vspace{2mm}
\textbf{tools-log:}
\begin{lstlisting}[language=json]
[{
    "tool": "search",
    "args": {
        "query": ["\"A Lifestyle Gap\" PhD thesis", "\"A Lifestyle Gap\" Lisa Portolan", "\"Slow Love\" podcast Lisa Portolan film event", "Lisa Portolan podcast \"met at a film event\"", "\"my committed relationship came to an end\" podcast"]
    },
    "status": "success"
}]
\end{lstlisting}

\textbf{summary:} Search results strongly suggest the author is Lisa Portolan. The text 'my committed relationship came to an end and i started a podcast with whom i first met at a film event' appears in snippets linked to her.

\vspace{0.2cm}

\noindent\colorbox{gray!70}{%
\parbox{\dimexpr\textwidth-2\fboxsep\relax}{%
\centering
\textbf{[Memory Unit 5]: Round 7}%
}%
}

\vspace{0.2cm}

\textbf{sub-goal:} Confirm Lisa Portolan as the author, verify the thesis title, and identify the podcast name

\vspace{2mm}
\textbf{tools-log:}
\begin{lstlisting}[language=json]
[{
    "tool": "search",
    "args": {
        "query": ["\"A Lifestyle Gap\" thesis", "\"A Lifestyle Gap\" dating apps", "\"started a podcast\" \"film event\" thesis", "site:edu \"dedicated to my children\" \"committed relationship\" podcast", "\"my committed relationship came to an end\" thesis acknowledgment"]
    },
    "status": "success"
}]
\end{lstlisting}

\textbf{summary:} Scraper results contain the mashed-up snippet linking 'A Lifestyle Gap', 'dedicated to my children', and the 'film event' story. The author is likely Lisa Portolan (Western Sydney University, est. 1989).

\vspace{0.2cm}

\noindent\colorbox{gray!70}{%
\parbox{\dimexpr\textwidth-2\fboxsep\relax}{%
\centering
\textbf{[Memory Unit 6]: Round 8\textasciitilde9}%
}%
}

\vspace{0.2cm}

\textbf{sub-goal:} Confirm the podcast name (likely 'Slow Love') by verifying the 'met at a film event' story in Lisa Portolan's thesis acknowledgments

\vspace{2mm}
\textbf{tools-log:}
\begin{lstlisting}[language=json]
[   
    {
        "tool": "search",
        "args": {
            "query": ["Lisa Portolan thesis acknowledgment \"film event\" podcast", "\"Slow Love\" podcast Lisa Portolan co-host", "Lisa Portolan \"committed relationship came to an end\"", "Did Lisa Portolan start a podcast with someone she met at a film event?", "thesis dedicated to my children dating apps podcast film event"]
        }",
        "status": "success"
    },
    {
        "tool": "search",
        "args": {
            "query": ["Lisa Portolan Jodi McAlister podcast", "Lisa Portolan \"Slow Love\" podcast co-host", "\"Slow Love\" podcast Lisa Portolan met at a film event", "Lisa Portolan podcast \"committed relationship came to an end\"", "Lisa Portolan thesis acknowledgements text", "Lisa Portolan \"The 40 Year Old Wino\"", "Lisa Portolan \"Dating in the Digital Age\" podcast"]
        },
        "status": "success"
    }
] 
\end{lstlisting}

\textbf{summary:} Lisa Portolan is confirmed as the author. Search results explicitly link the phrase 'met at a film event' and 'started a podcast' to her and the podcast 'Slow Love'.

\vspace{0.2cm}

\noindent\colorbox{gray!70}{%
\parbox{\dimexpr\textwidth-2\fboxsep\relax}{%
\centering
\textbf{[Memory Unit 7]: Round 10}%
}%
}

\vspace{0.2cm}

\textbf{sub-goal:} Verify the podcast name and the co-host meeting story within Lisa Portolan's thesis acknowledgments

\vspace{2mm}
\textbf{tools-log:}
\begin{lstlisting}[language=json]
[
    {
        "tool": "search",
        "args": {
            "query": ["Lisa Portolan thesis acknowledgment \"committed relationship came to an end\"", "\"Slow Love\" podcast Lisa Portolan started with", "Lisa Portolan podcast \"met at a film event\"", "Did Lisa Portolan meet her podcast co-host at a film event?", "Lisa Portolan thesis \"dedicated to my children\""]
        },
        "status": "success"
    }
]
\end{lstlisting}

\textbf{summary:} The snippet 'Slow Love Podcast Lisa Portolan Co-host Met At Film Event' appearing in SEO results strongly suggests 'Slow Love' is the correct answer.

\vspace{0.2cm}

\noindent\colorbox{gray!70}{%
\parbox{\dimexpr\textwidth-2\fboxsep\relax}{%
\centering
\textbf{[Memory Unit 8]: Round 11\textasciitilde12}%
}%
}

\vspace{0.2cm}

\textbf{sub-goal:} Identify the podcast name and author by verifying the specific acknowledgment phrase

\vspace{2mm}
\textbf{tools-log:}
\begin{lstlisting}[language=json]
[
    {
        "tool": "search",
        "args": {
            "query": ["\"started a podcast with whom I first met at a film event\"", "\"my committed relationship came to an end\" thesis", "\"A Lifestyle Gap\" PhD thesis dating apps", "Lisa Portolan thesis \"film event\"", "podcast co-host \"met at a film event\""]
        },
        "status": "success"
    },
    {
        "tool": "google_scholar",
        "args": {
            "query": ["\"A Lifestyle Gap\" PhD thesis", "\"A Lifestyle Gap\" dating apps", "\"started a podcast\" \"film event\" thesis", "site:edu \"dedicated to my children\" \"committed relationship\" podcast", "\"my committed relationship came to an end\" thesis acknowledgment"]
        },
        "status": "success"
    }
]
\end{lstlisting}

\textbf{summary:} Google Scholar returned no results for 'A Lifestyle Gap', indicating it's likely a scraper artifact. Lisa Portolan remains the primary candidate (WSU est. 1989, dating apps topic, 'Slow Love' podcast).

\end{tcolorbox}

\paragraph{Example 2: Framework}

This example demonstrates the framework's capability to handle complex real-world from the GAIA benchmark. The trajectory highlights the seamless orchestration between specialized sub-agents, facilitated by structured memory management: the Browser-Use Agent precisely locates the dataset URL, while the Data Analysis Agent downloads the CSV file and executes Python code to perform rigorous data cleaning and verification, overcoming the calculation errors and context limitations common in standard LLMs.

\begin{tcolorbox}[breakable,title=Case Trajectory for Example 2 in GAIA]
\textbf{Question:} According to the World Bank, which countries had gross savings of over 35\% of GDP for every year in the period 2001-2010? Give your answer as a comma-separated list of countries in alphabetical order. Use the countries most common names in english when answering.

\vspace{0.2cm}

\noindent\colorbox{gray!70}{%
    \parbox{\dimexpr\textwidth-2\fboxsep\relax}{%
    \centering
        \textbf{Step 1: Planning \& Information Seeking}%
    }%
}

\vspace{0.2cm}

\textcolor{blue1}{\textbf{<think>} \\
We need to find data from World Bank on gross savings (\% of GDP) for each country for each year 2001-2010 inclusive. Then find those countries where gross savings >35\% for each year in that period. \\
\textbf{</think>}}

\vspace{0.2cm}

\textbf{<tool\_call> \\ 
\texttt{\{"name": "search", "arguments": \{"query": ["World Bank gross savings \% of GDP by country 2001-2010 data"]\}\}}
\\
\textbf{</tool\_call>}}

\vspace{0.2cm}

\textbf{<tool\_response>}\\
A Google search for `World Bank gross savings \% of GDP by country 2001-2010 data' found 10 results: \\[0.1cm]
1. [Gross savings (\% of GDP)] (https://data.worldbank.org/indicator/NY.GNS.ICTR.ZS) \\ 
Gross savings (\% of GDP) Country official statistics, National Statistical Organizations and/or Central Banks; National Accounts data files, Organisation for ... \\[0.1cm]
...\\
\textbf{</tool\_response>}

\vspace{0.2cm}

\textcolor{green1}{\textbf{<memory>} \\
steps [1]: \\
\{``sub\_goal'': ``Identify the World Bank indicator code for 'Gross savings (\% of GDP)'', ``tools\_log'': ..., ``summary'': ``The agent searched for World Bank gross savings data and located the official indicator page, identifying the relevant indicator code as `NY.GNS.ICTR.ZS'.''\}\\
\textbf{</memory>}
}

\colorbox{gray!70}{%
    \parbox{\dimexpr\textwidth-2\fboxsep}{%
    \centering
        \textbf{Step 2: Navigation \& Extraction (Browser-Use Agent)}%
    }%
}

\vspace{0.2cm}

\textcolor{blue1}{\textbf{<think>} \\
We need to find the list of countries where gross savings (\% of GDP) >35\% for each year 2001-2010 inclusive. The World Bank indicator for gross savings (\% of GDP) is "NY.GNS.ICTR.ZS". Let's click that. \\
\textbf{</think>}
}

\vspace{0.2cm}

\textcolor{black}{\textbf{<tool\_call>} \\ 
\texttt{\{"name": "visit", "arguments": \{"url":  ["https://data.worldbank.org/indicator/\\NY.GNS.ICTR.ZS"], "goal": "Find the data for gross savings (\% of GDP) for all countries for years 2001-2010 and identify which countries have values >35\% for each year."\}\}}
\\
\textbf{</tool\_call>}}

\vspace{0.2cm}

\noindent\hspace{0.5cm}\textcolor{gray1}{\textit{$\hookrightarrow$ Delegating task to \textbf{Browser-Use Agent}...}}

\vspace{0.1cm}

\textbf{<tool\_response>}\\
Screenshot:\\
\begin{center}
  \includegraphics[width=0.6\linewidth]{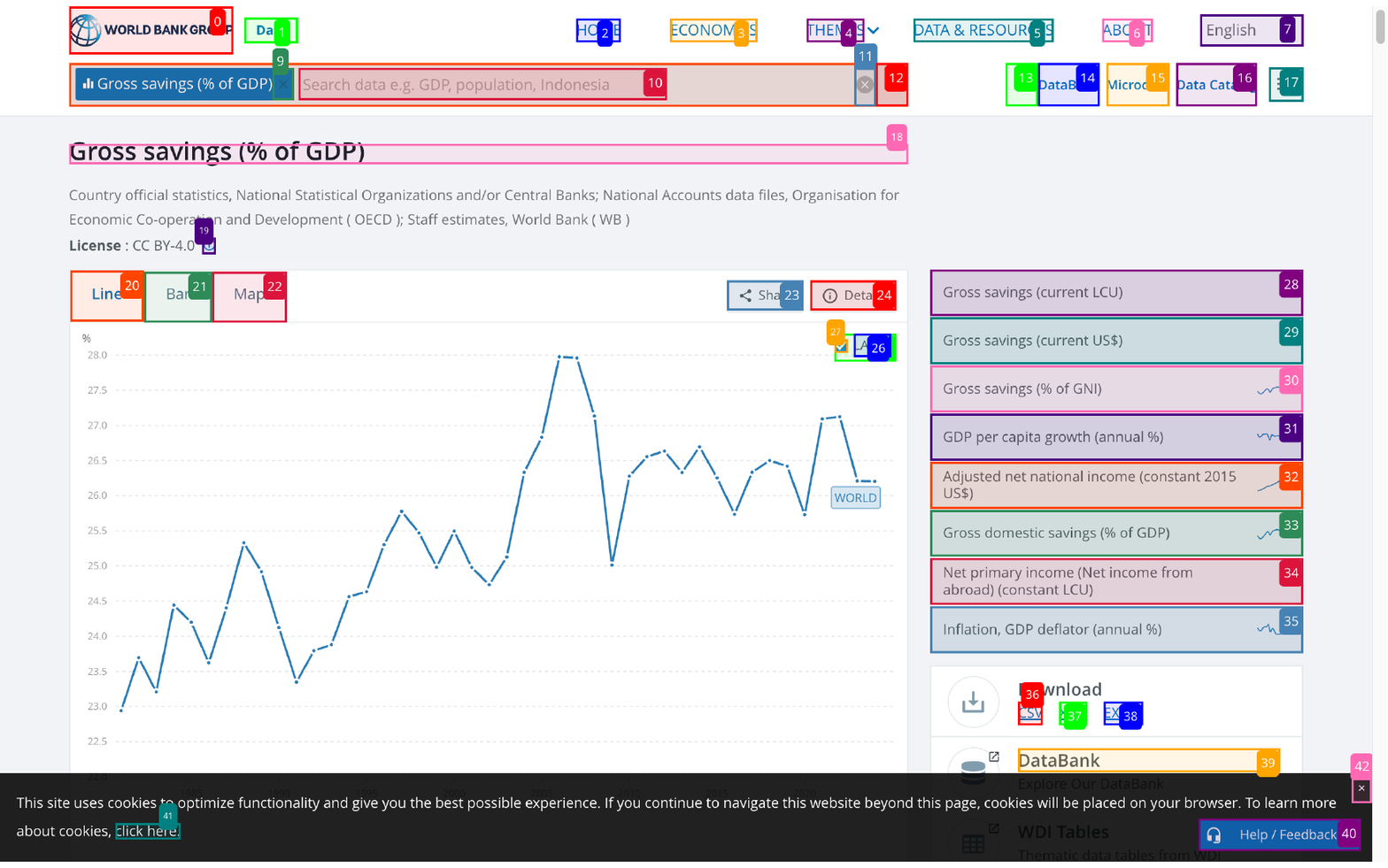}
\end{center}
Thinking trace:\\
I see the download options on the right side of the page under the ``Download'' header. There are links for CSV, XML, and EXCEL. I need to get the direct download link for the CSV or Excel file. I will check the `href` attribute of element [36] (CSV) which is `https://api.worldbank.org/v2/en/indicator/NY.GNS.ICTR.ZS?\\downloadformat=csv'. This looks like a direct download link. I will provide this link. \\[0.1cm]
Summary: \\
The World Bank Open Data platform provides comprehensive data on `Gross savings (\% of GDP)' (indicator NY.GNS.ICTR.ZS) for all countries from 1960 to 2024, which includes the requested time period of 2001–2010. The data can be downloaded in CSV formats via the provided link (https://api.worldbank.org/v2/en/indicator/NY.GNS.ICTR.ZS?downloadformat=csv), enabling users to extract and analyze values for each country. \\
\textbf{</tool\_response>}

\vspace{0.2cm}

\colorbox{gray!70}{%
    \parbox{\dimexpr\textwidth-2\fboxsep}{%
    \centering
        \textbf{Step 3: Data Processing (Data Analysis Agent)}%
    }%
}

\vspace{0.2cm}

\textcolor{blue1}{\textbf{<think>} \\
The page indicates we can download CSV. Let's download the CSV file. \\
\textbf{</think>}
}

\vspace{0.2cm}

\textcolor{black}{\textbf{<tool\_call>} \\ 
\texttt{\{"name": "visit", "arguments": \{"url":  ["https://api.worldbank.org/v2/en/indicator/\\NY.GNS.ICTR.ZS?downloadformat=csv"], "goal": "Download the CSV file containing gross savings (\% of GDP) data for all countries from 1960 to 2024."\}\}}
\\
\textbf{</tool\_call>}}

\vspace{0.2cm}

\noindent\hspace{0.5cm}\textcolor{gray1}{\textit{$\hookrightarrow$ Delegating task to \textbf{Data Analysis Agent}...}}

\vspace{0.1cm}

\textbf{<tool\_response>}\\
Deep analysis for extracted file Metadata\_Indicator\_API\_NY.GNS.ICTR.ZS\_DS2\_en\_csv\_v2\_2962.csv in zip file:\\
<rational>\\
The provided file, `Metadata\_Indicator\_API\_NY.GNS.ICTR.ZS\_DS2\_en\_csv\_v2\_2962.csv', is a metadata file documenting the definition and source of the indicator ``Gross savings (\% of GDP)''. It contains descriptive text (e.g., ``Savings is an amount that represent...'') but does not contain any numerical time-series data or country-specific rows. To answer the ultimate question regarding which countries had gross savings of over 35\% of GDP between 2001 and 2010, the primary data file (typically named `API\_NY.GNS.ICTR.ZS\_DS2\_en\_csv\_v2\_... .csv' in World Bank datasets) containing the actual values for countries and years is required. This file is irrelevant for the calculation step.\\
</rational>\\[0.1cm]
=======\\[0.1cm]
Deep analysis for extracted file API\_NY.GNS.ICTR.ZS\_DS2\_en\_csv\_v2\_2962.csv in zip file:\\
<rational>\\
The user's goal is to identify countries with gross savings (\% of GDP) strictly greater than 35\% for every year in the period 2001-2010 using the provided CSV file.\\
I executed a Python script (``Path B'') to process the file `API\_NY.GNS.ICTR.ZS\_DS2\_en\_csv\_v2\_2962.csv'. The script:\\
1.  Read the CSV file, correctly skipping the metadata rows.\\
2.  Filtered out non-country regional aggregates (e.g., ``Arab World'', ``High income'') to ensure only sovereign countries were listed.\\
3.  Checked the condition `Gross savings > 35.0' for each country for every year from 2001 through 2010.\\
4.  Renamed countries to their common English names (e.g., ``Brunei Darussalam'' to ``Brunei'').\\
5.  Output the resulting list alphabetically.\\
The code execution was successful and returned the list: `Brunei, China, Morocco, Singapore'. These are the countries that met the specified criteria according to the data in the provided file.\\
</rational>\\[0.1cm]
<evidence>\\
The code execution produced the following output: `Brunei, China, Morocco, Singapore'\\
</evidence>\\[0.1cm]
<summary>\\
According to the World Bank data in the provided file, the countries that had gross savings of over 35\% of GDP for every year in the period 2001-2010 are Brunei, China, Morocco, and Singapore.\\
</summary>\\[0.1cm]
=======\\[0.1cm]
Deep analysis for extracted file Metadata\_Country\_API\_NY.GNS.ICTR.ZS\_DS2\_en\_csv\_v2\_2962.csv in zip file:\\
<rational>\\
The provided file `Metadata\_Country\_API\_NY.GNS.ICTR.ZS\_DS2\_en\_csv\_v2\_2962.csv' contains metadata regarding countries and regional aggregates (e.g., Region, IncomeGroup, SpecialNotes) but does not contain the actual Gross Savings (\% of GDP) time-series data. The values required to determine which countries had gross savings over 35\% of GDP from 2001 to 2010 are missing from this file. The user likely intended to provide the primary data file (usually named `API\_NY.GNS.ICTR.ZS\_DS2\_en\_csv\_v2\_2962.csv'). Consequently, this file is irrelevant for performing the specific data analysis requested in the ultimate question, although it would be useful for filtering out non-country aggregates if the data file were present.\\
</rational>\\
\textbf{</tool\_response>}

\vspace{0.2cm}

\textcolor{green1}{\textbf{<memory>} \\
steps [2, 3]: \\
\{``sub\_goal'': ``Download and process the 'Gross savings (\% of GDP)' dataset'', ``tools\_log'': ..., ``summary'': ``The agent visited the World Bank indicator page, confirmed the code `NY.GNS.ICTR.ZS', and identified the CSV download URL. Subsequently, the agent successfully executed a self-contained Python script to download, extract, and process the World Bank dataset directly. The script filtered for entities with gross savings > 35\% for every year from 2001 to 2010. The resulting list includes: Brunei, China, Morocco, Singapore. The agent is now ready to provide the final answer.''\}\\
\textbf{</memory>}
}

\vspace{0.2cm}

\colorbox{gray!70}{%
    \parbox{\dimexpr\textwidth-2\fboxsep}{%
    \centering
        \textbf{Step 4: Final Synthesis}%
    }%
}

\vspace{0.2cm}

\textcolor{blue1}{\textbf{<think>} \\
I will output the answer as a comma-separated list of common names. Common names: Brunei, China, Morocco, Singapore. One final check: Is there any other country? My script filtered all countries in the CSV. The CSV contains all World Bank data. I am confident in the list: Brunei, China, Morocco, Singapore. Final Answer seems ready. \\
\textbf{</think>}
}

\vspace{0.2cm}

\textbf{<answer>}\\
Brunei, China, Morocco, Singapore \\
\textbf{</answer>}

\end{tcolorbox}

\section*{B $\;$ Evaluation Details}\label{append:bench}

For the GAIA benchmark, we employ Qwen3-235b-a22b-2507 as the judge model, utilizing the text-only validation subset in accordance with the settings in \cite{li2025webthinker}. For Humanity's Last Exam (HLE), we evaluate performance on the subset of 2,158 text-only questions. For BrowseComp, BrowseComp-ZH, and HLE, we utilize GPT-4o-2024-08-06 as the judge model. To ensure consistency and comparability, all evaluation prompts across these benchmarks adhere to their official protocols. We selected these four benchmarks to provide a comprehensive assessment of our framework's capabilities:

\begin{itemize}
    \item \textbf{GAIA\cite{mialon2023gaia}} - A benchmark comprising real-world, challenging tasks designed to evaluate general AI assistants. We utilize the 103-sample text-only subset to assess capabilities such as multi-step reasoning, web browsing, and general tool-use proficiency.
    \item \textbf{BrowseComp\cite{wei2025browsecomp}} - A simple yet challenging benchmark consisting of 1,266 questions designed to measure an agent's web browsing capabilities. It requires agents to persistently navigate the internet to locate hard-to-find and entangled information.
    \item \textbf{BrowseComp-ZH\cite{zhou2025browsecompzh}} - A high-difficulty Chinese benchmark containing 289 multi-hop questions specifically tailored to retrieve information within the Chinese web ecosystem.
    \item \textbf{HLE\cite{phan2025humanitys}} - Humanity's Last Exam is an expert-curated benchmark featuring challenging questions that span a wide range of disciplines. It is designed to test expert-level performance on closed-ended, verifiable questions involving cutting-edge scientific knowledge.
\end{itemize}

\section*{C $\;$ Details of Basic Tools}
The framework is equipped with a core set of fundamental capabilities, serving as the primitives for both the Main Agent and the specialized sub-agents. 


\paragraph{Search.}
To acquire external knowledge, the \texttt{Search} tool interfaces with search engines (e.g., Google, Google Scholar). It retrieves relevant URLs and snippets, enabling the agent to locate information sources dynamically.

\paragraph{Read \& Parse.}
Information ingestion is handled by a hybrid strategy tailored to the source format: \textit{(i) Web Content:} For web pages, the tool utilizes lightweight APIs (e.g., Jina Reader) to convert HTML into clean Markdown. This serves as a lightweight, high-efficiency complement to the Browser-Use GUI Agent. \textit{(ii) File Content:} A unified \texttt{FileParser} handles diverse local formats.

\paragraph{Code Execution.}
For tasks requiring computation or logic verification, the \texttt{Code Execution} tool provides a secure, Docker-based Python sandbox. It supports file I/O and library imports, allowing the agent to solve mathematical problems, process data, and verify algorithmic logic in an isolated environment.


\newpage
\bibliographystyle{unsrt}  
\bibliography{references}

\end{document}